\PassOptionsToPackage{unicode}{hyperref}
\PassOptionsToPackage{hyphens}{url}
\PassOptionsToPackage{dvipsnames,svgnames*,x11names*}{xcolor}
\documentclass[
  11pt,
  letterpaper,
]{article}
\usepackage{lmodern}
\usepackage{amssymb,amsmath}
\usepackage{ifxetex,ifluatex}
\ifnum 0\ifxetex 1\fi\ifluatex 1\fi=0 % if pdftex
  \usepackage[T1]{fontenc}
  \usepackage[utf8]{inputenc}
  \usepackage{textcomp} % provide euro and other symbols
\else % if luatex or xetex
  \usepackage{unicode-math}
  \defaultfontfeatures{Scale=MatchLowercase}
  \defaultfontfeatures[\rmfamily]{Ligatures=TeX,Scale=1}
\fi
\IfFileExists{upquote.sty}{\usepackage{upquote}}{}
\IfFileExists{microtype.sty}{% use microtype if available
  \usepackage[]{microtype}
  \UseMicrotypeSet[protrusion]{basicmath} % disable protrusion for tt fonts
}{}
\makeatletter
\@ifundefined{KOMAClassName}{% if non-KOMA class
  \IfFileExists{parskip.sty}{%
    \usepackage{parskip}
  }{% else
    \setlength{\parindent}{0pt}
    \setlength{\parskip}{6pt plus 2pt minus 1pt}}
}{% if KOMA class
  \KOMAoptions{parskip=half}}
\makeatother
\usepackage{xcolor}
\IfFileExists{xurl.sty}{\usepackage{xurl}}{} % add URL line breaks if available
\IfFileExists{bookmark.sty}{\usepackage{bookmark}}{\usepackage{hyperref}}
\hypersetup{
  pdftitle={When Single-Dataset Conclusions Fail: A 45-Task Study of Threshold Tuning and Resampling for Imbalanced Classification},
  pdfauthor={Diyorbek Musayev; Queensland University of Technology, Brisbane, Australia; diyorbek.musaev@connect.qut.edu.au; ORCID 0009-0005-2249-0834},
  colorlinks=true,
  linkcolor=blue,
  filecolor=Maroon,
  citecolor=Blue,
  urlcolor=blue,
  pdfcreator={LaTeX via pandoc}}
\usepackage[margin=1in]{geometry}
\usepackage{longtable,booktabs}
\usepackage{etoolbox}
\makeatletter
\patchcmd\longtable{\par}{\if@noskipsec\mbox{}\fi\par}{}{}
\makeatother
\IfFileExists{footnotehyper.sty}{\usepackage{footnotehyper}}{\usepackage{footnote}}
\makesavenoteenv{longtable}
\usepackage{graphicx}
\makeatletter
\def\maxwidth{\ifdim\Gin@nat@width>\linewidth\linewidth\else\Gin@nat@width\fi}
\def\maxheight{\ifdim\Gin@nat@height>\textheight\textheight\else\Gin@nat@height\fi}
\makeatother
\setkeys{Gin}{width=\maxwidth,height=\maxheight,keepaspectratio}
\makeatletter
\def\fps@figure{htbp}
\makeatother
\providecommand{\tightlist}{%
  \setlength{\itemsep}{0pt}\setlength{\parskip}{0pt}}
\title{When Single-Dataset Conclusions Fail: A 45-Task Study of
Threshold Tuning and Resampling for Imbalanced Classification}
\author{Diyorbek Musayev \and Queensland University of Technology,
Brisbane, Australia \and diyorbek.musaev@connect.qut.edu.au \and ORCID
0009-0005-2249-0834}
\date{Preprint, August 2026}

\begin{document}
\maketitle

\hypertarget{abstract}{%
\section{Abstract}\label{abstract}}

Class-imbalance handling for binary classification is routinely
evaluated on a single benchmark dataset, and conclusions drawn from that
dataset are reported as if they were properties of the method. We show
concretely that this practice is unsafe. Starting from the public Kaggle
credit-card fraud dataset --- among the most heavily benchmarked
imbalanced-classification tasks in the literature --- we establish under
a leakage-free nested cross-validation protocol that a plain Random
Forest at the default 0.5 decision threshold attains F1 = 0.861 $\pm$ 0.021,
and that decision-threshold tuning yields it no benefit whatsoever ($\Delta$F1
= -0.002). Read in isolation, this supports an appealing conclusion: for
a well-calibrated ensemble, the entire apparatus of imbalance handling
is unnecessary.

We then apply the identical protocol to a suite of 45 binary tasks
spanning imbalance ratios from 1:1.5 to 1:178, comprising 2,025 model
fits across four model families. The conclusion reverses. Random Forest
is the model family that benefits \emph{most} from threshold tuning
across the suite ($\Delta$F1 = +0.101 $\pm$ 0.134), not least; three other families
replicate their fraud-dataset behaviour almost exactly, isolating Random
Forest on the fraud dataset as the anomaly. Similarly, Synthetic
Minority Over-sampling (SMOTE) is harmful on the fraud dataset but
beneficial across the suite (mean $\Delta$F1 = +0.076, 138 wins / 39 losses,
Wilcoxon \emph{p} = $2.7 \times 10^{-17}$).

We report two further results. First, the benefit of threshold tuning is
non-monotonic in the imbalance ratio, near zero below 1:5, peaking at
$\Delta$F1 = +0.120 in the 1:15--1:40 band, and declining to +0.045 beyond
1:100 --- which explains why the fraud dataset, at 1:577, is an
unrepresentative place to study the question. Second, we test and reject
an intuitive heuristic: validation-set calibration error does not
predict how much threshold tuning will help (expected calibration error
\emph{r} = -0.087; Brier score \emph{r} = +0.137), so practitioners
cannot use calibration diagnostics to decide whether tuning is
worthwhile. We release the protocol, the 45-task harness, and all
per-run metrics.

\textbf{Keywords:} class imbalance; decision-threshold tuning; SMOTE;
external validity; generalisation of empirical findings; credit-card
fraud detection; reproducible evaluation.

\begin{center}\rule{0.5\linewidth}{0.5pt}\end{center}

\hypertarget{introduction}{%
\section{1. Introduction}\label{introduction}}

A large empirical literature studies how to train binary classifiers
when the positive class is rare. The standard interventions ---
resampling the training set, weighting the loss by inverse class
frequency, and moving the decision threshold away from 0.5 --- are well
established, and a steady stream of papers compares them. A striking
feature of this literature is how often the comparison is conducted on a
\emph{single} dataset, with the resulting ranking of interventions
reported as a general finding.

This paper asks whether that inference is safe, and answers no.

Our entry point is the public Kaggle credit-card fraud dataset of Dal
Pozzolo et al., which contains 284,807 transactions of which 492 (0.172
\%) are fraudulent. It is among the most heavily used
imbalanced-classification benchmarks in existence; hundreds of papers
report results on it. Reported F1 scores on the fraud class span an
extraordinary range, from roughly 0.70 for unaided classical baselines
to above 0.98 for some recent deep architectures. Recent methodological
work attributes a substantial part of this spread not to algorithmic
progress but to three specific evaluation flaws: (i) \emph{preprocessing
leakage}, where a scaler or oversampler is fitted on data including the
test split; (ii) \emph{evaluation-set threshold selection}, where the
decision threshold is chosen to maximise the very metric being reported
on the very split being reported; and (iii) \emph{inadequate temporal
validation}, where random rather than time-ordered folds are used
despite a time axis being present (Hayat \& Magnier, 2025; Kabane,
2024).

We first eliminate flaws (i) and (ii) by specifying a nested
cross-validation protocol in which the decision threshold is selected on
an inner validation fold disjoint from both the training data and the
reporting test fold. Applying this protocol to five model families on
the fraud dataset yields a clean and quotable result: a plain Random
Forest at the default threshold attains F1 = 0.861 $\pm$ 0.021, and
threshold tuning changes nothing ($\Delta$F1 = -0.002). SMOTE makes it worse.
The natural conclusion --- that a well-calibrated ensemble needs no
imbalance handling at all --- is exactly the kind of claim this
literature routinely makes on the strength of one dataset.

We then test that conclusion. Using the same protocol, the same four
supervised model families, and the same four intervention strategies, we
evaluate 45 binary classification tasks with imbalance ratios from 1:1.5
to 1:178 --- 2,025 model fits in total. Three of the four model families
behave on the suite almost exactly as they behaved on the fraud dataset.
Random Forest does not: across the suite it gains more from threshold
tuning than any other family ($\Delta$F1 = +0.101), where on the fraud dataset
it gained nothing. The single-dataset conclusion is not merely
imprecise; on the specific point that made it interesting, it is
backwards.

The contributions of this paper are as follows.

\begin{enumerate}
\def\labelenumi{\arabic{enumi}.}
\item
  \textbf{A concrete demonstration that single-dataset
  imbalance-handling conclusions fail to generalise}, on the
  most-benchmarked dataset in this application area, with the failure
  localised to a specific model family and quantified against a 45-task
  reference. We are not aware of prior work that demonstrates this
  reversal directly rather than arguing for it in principle.
\item
  \textbf{A characterisation of what does generalise.} Threshold tuning
  helps on average (mean $\Delta$F1 = +0.059 across dataset $\times$ model pairs; 138
  wins, 31 losses; Wilcoxon \emph{p} = $4.1 \times 10^{-20}$) and SMOTE helps on
  average (+0.076; 138 wins, 39 losses; \emph{p} = $2.7 \times 10^{-17}$), but the
  magnitude is strongly model-dependent: SMOTE is worth +0.170 F1 to an
  MLP and +0.005 --- nothing --- to logistic regression.
\item
  \textbf{An inverted-U relationship between imbalance ratio and
  threshold-tuning benefit}, with the peak at moderate imbalance
  (1:15--1:40, $\Delta$F1 = +0.120) and decline at both mild (\textless1:5, $\Delta$F1
  = -0.009) and extreme (\textgreater1:100, $\Delta$F1 = +0.045) imbalance.
  This is not the monotone relationship one would naively assume, and it
  identifies extreme-imbalance datasets such as credit-card fraud as
  poor venues for studying threshold selection.
\item
  \textbf{A negative result of practical consequence.} It is intuitive
  that threshold tuning should help in proportion to how miscalibrated a
  model's scores are, which would give practitioners a cheap diagnostic:
  measure calibration error on validation, and tune only when it is
  high. We test this and reject it. Expected calibration error is
  uncorrelated with tuning benefit (\emph{r} = -0.087), and the Brier
  score is only weakly correlated (\emph{r} = +0.137). The diagnostic
  does not work.
\end{enumerate}

Section 2 reviews prior work, distinguishing the substantial literature
that establishes \emph{that} threshold tuning helps from the question we
address, which is \emph{whether single-dataset findings about it
transfer}. Section 3 specifies the protocol and both dataset suites.
Section 4 reports the credit-card fraud study. Section 5 reports the
45-task study and the reversal. Section 6 analyses what generalises.
Section 7 discusses limitations --- of which the composition of our
dataset suite is the most serious --- and concludes.

\begin{center}\rule{0.5\linewidth}{0.5pt}\end{center}

\hypertarget{related-work}{%
\section{2. Related Work}\label{related-work}}

\hypertarget{threshold-moving-and-resampling}{%
\subsection{2.1 Threshold moving and
resampling}\label{threshold-moving-and-resampling}}

Adjusting the decision threshold of a probabilistic classifier is a
long-established response to class imbalance, traceable to work on
cost-sensitive classification and ROC-based operating-point selection
(Provost \& Fawcett, 2001; Sheng \& Ling, 2006). Resampling approaches,
of which SMOTE (Chawla et al., 2002) is the most widely used, instead
alter the training distribution. Cost-sensitive weighting forms a third
family. Standard references treat these as alternative means to the same
end.

Critically for our purposes, the claim that threshold tuning is
\emph{broadly effective} is already well supported by multi-dataset
evidence, and we do not claim it as a contribution. Esposito et
al.~(2021) introduce GHOST, an automated threshold-selection procedure,
and validate it on 138 public drug-discovery datasets, reporting that
most classifiers benefit and that Random Forest benefits substantially.
M-Tune (Molecular Diversity) reports comparable findings on a similar
scale. ``Balancing the Scales'' (arXiv:2409.19751, 2024) compares SMOTE,
class weighting, and decision-threshold calibration across multiple
datasets and models and concludes that threshold calibration is the most
consistently effective of the three. The scikit-learn library ships
\texttt{TunedThresholdClassifierCV} for cross-validated threshold
selection, with explicit documentation warning against selecting the
threshold on data also used for fitting.

Our contribution is therefore \emph{not} that threshold tuning helps. It
is that conclusions of the form ``intervention X is or is not needed for
model Y'', when drawn from one dataset, do not transfer --- and we
demonstrate a case where such a conclusion inverts.

\hypertarget{the-credit-card-fraud-benchmark}{%
\subsection{2.2 The credit-card fraud
benchmark}\label{the-credit-card-fraud-benchmark}}

The Kaggle credit-card dataset originates with Dal Pozzolo's work on
adaptive fraud detection (Dal Pozzolo, 2015). Its features V1--V28 are
PCA projections applied for privacy; only \texttt{Time} and
\texttt{Amount} are interpretable. Dal Pozzolo et al.~(2015)
additionally observed that undersampling distorts posterior
probabilities and requires calibration correction --- an early
indication that resampling and threshold choice interact.

A large secondary literature uses the dataset as a testbed. Popova \&
Gardi (arXiv:2509.15044, 2025) benchmark five classifiers under
undersampling, SMOTE, and hybrid regimes. Stacking ensembles (IJACSA
Vol. 15 No.~10, 2024) report F1 = 0.87. Deep approaches include
heterogeneous graph autoencoders (Singh et al., arXiv:2410.08121, 2024;
F1 = 0.81, AUPRC = 0.89), GAT+VAE ensembles (MDPI, 2025; F1
\textgreater{} 0.98), adversarial autoencoders, and
transformer-conditioned GAN oversampling (arXiv:2509.19032, 2025).
Threshold treatment is unreported in essentially all of them, which ---
since every mainstream library defaults to 0.5 in \texttt{.predict()}
--- should be read as the default threshold having been used.

\hypertarget{methodological-critiques-and-external-validity}{%
\subsection{2.3 Methodological critiques and external
validity}\label{methodological-critiques-and-external-validity}}

Hayat \& Magnier (arXiv:2506.02703, 2025) document leakage, vague
reporting, inadequate temporal validation, and recall-optimising metric
manipulation in this literature, demonstrating that a minimal network
with deliberate leakage reaches 99.9 \% recall. Kabane
(arXiv:2412.07437, 2024) shows in controlled fashion that applying
sampling before the train/test split inflates XGBoost results.

These papers concern \emph{internal} validity --- whether a reported
number is a faithful estimate of the model's performance on its own
dataset. The present paper concerns \emph{external} validity --- whether
a faithful, leakage-free finding on one dataset licenses a general
claim. The two are complementary: we adopt their protocol
recommendations in full, and then show that even a methodologically
clean single-dataset result can mislead.

\begin{center}\rule{0.5\linewidth}{0.5pt}\end{center}

\hypertarget{methods}{%
\section{3. Methods}\label{methods}}

\hypertarget{evaluation-protocol}{%
\subsection{3.1 Evaluation protocol}\label{evaluation-protocol}}

Both studies use one protocol. For each (dataset, model, strategy)
combination:

\begin{enumerate}
\def\labelenumi{\arabic{enumi}.}
\item
  \textbf{Outer split.} Stratified \emph{k}-fold partition of the
  dataset (\emph{k} = 5 for the fraud study, \emph{k} = 3 for the
  45-task suite, the reduction being a compute-budget concession). Fold
  \emph{k} is the test fold; the remainder is the outer training set.
\item
  \textbf{Inner split.} A stratified split of the outer training set
  into an inner training set and an inner validation set (80/20 for the
  fraud study, 75/25 for the suite). The validation set is used
  \emph{only} to select the decision threshold and never to fit model
  parameters.
\item
  \textbf{Preprocessing.} \texttt{StandardScaler} is fitted on the inner
  training set alone and applied to the validation and test folds. Where
  SMOTE is used it is applied only to the inner training set, after
  scaling. Neither the validation nor the test fold is resampled.
\item
  \textbf{Threshold selection.} For the \texttt{plain} strategy the
  threshold is fixed at 0.5. Otherwise the threshold is *t** = argmax
  over the grid \{0.01, 0.02, \ldots, 0.99\} of F1 on the inner
  validation fold.
\item
  \textbf{Reporting.} *t** is applied to the test fold. We record
  precision, recall, F1, ROC-AUC, AUPRC, expected calibration error
  (both equal-width and equal-frequency binning), and the Brier score.
  Calibration statistics are computed on the \emph{validation} fold,
  since that is the information available to a practitioner deciding
  whether to tune.
\end{enumerate}

No information from the test fold influences preprocessing, fitting, or
threshold selection.

\hypertarget{the-paired-f1-difference-statistic}{%
\subsection{3.2 The paired F1-difference
statistic}\label{the-paired-f1-difference-statistic}}

Our central quantity is
\[\Delta F1 = F1(\text{threshold } t^*) - F1(\text{threshold } 0.5),\]
computed on the same test fold using the same fitted model. It is
therefore a paired within-fold contrast, isolating the effect of the
threshold choice and eliminating between-fold and between-model
variance.

\hypertarget{model-families-and-strategies}{%
\subsection{3.3 Model families and
strategies}\label{model-families-and-strategies}}

Four supervised families are common to both studies: \textbf{Logistic
Regression} (L-BFGS, \texttt{max\_iter} 500--1000); \textbf{Random
Forest} (100 trees, \texttt{max\_depth} 16); \textbf{Histogram Gradient
Boosting} (\texttt{max\_iter} 100--200, learning rate 0.1); and a
\textbf{Multi-Layer Perceptron} (hidden layers 64--32, ReLU, Adam, early
stopping). The fraud study additionally includes an \textbf{Autoencoder}
anomaly detector trained on legitimate transactions only, scored by
reconstruction error. Hyperparameters are deliberately mainstream; the
object of study is the evaluation protocol and the interventions, not a
tuned model.

Four strategies are compared where well defined: \texttt{plain} (no
intervention, threshold 0.5); \texttt{tuned\_threshold} (threshold
selected on validation); \texttt{class\_balanced\_tuned}
(inverse-frequency class weights plus threshold selection); and
\texttt{smote\_tuned} (SMOTE on the inner training set plus threshold
selection). \texttt{class\_balanced\_tuned} is omitted for the MLP,
whose scikit-learn implementation accepts no class-weight argument.

\hypertarget{dataset-suites}{%
\subsection{3.4 Dataset suites}\label{dataset-suites}}

\textbf{Study A --- credit-card fraud.} The full Kaggle dataset (284,807
rows, 492 positives, imbalance 1:577), 5 outer folds, five model
families, 17 applicable (model, strategy) combinations, 85 fits.

\textbf{Study B --- 45 binary tasks.} A suite spanning imbalance ratios
1:1.5 to 1:178, comprising:

\begin{itemize}
\tightlist
\item
  \emph{Real, 14 tasks}: handwritten digits one-vs-rest (10 tasks,
  $\approx$1:9); breast cancer (1:1.7); wine one-vs-rest (3 tasks, 1:2--1:3).
\item
  \emph{Real, rarefied, 1 task}: breast cancer subsampled to 1:21.
\item
  \emph{Synthetic, 30 tasks}: \texttt{make\_classification} sweeping
  imbalance ratio (10, 30, 100, 300, 1000), class separation (0.6, 1.0,
  1.8), and dimensionality (20 and 50 features, 10 and 15 informative).
  The synthetic arm is the \emph{controlled} component of the study: it
  varies the imbalance ratio while holding the data-generating process
  fixed, which no collection of observational datasets can do.
\end{itemize}

Three folds per task, four model families, four strategies, giving 2,025
fits. Section 7.1 discusses the limitations of this composition frankly;
it is the weakest part of the study and the first thing a replication
should improve.

\begin{center}\rule{0.5\linewidth}{0.5pt}\end{center}

\hypertarget{study-a-the-credit-card-fraud-dataset}{%
\section{4. Study A: The Credit-Card Fraud
Dataset}\label{study-a-the-credit-card-fraud-dataset}}

Table 1 reports the five-fold results for the strongest configuration of
each model family on the fraud dataset.

\textbf{Table 1.} Credit-card fraud, 5-fold nested CV, mean $\pm$ standard
deviation. R@P$\geq$0.90 is the maximum recall attainable at precision at
least 0.90.

\begin{longtable}[]{@{}lllllll@{}}
\toprule
\begin{minipage}[b]{0.12\columnwidth}\raggedright
Model\strut
\end{minipage} & \begin{minipage}[b]{0.12\columnwidth}\raggedright
Strategy\strut
\end{minipage} & \begin{minipage}[b]{0.12\columnwidth}\raggedright
F1\strut
\end{minipage} & \begin{minipage}[b]{0.12\columnwidth}\raggedright
AUPRC\strut
\end{minipage} & \begin{minipage}[b]{0.12\columnwidth}\raggedright
ROC-AUC\strut
\end{minipage} & \begin{minipage}[b]{0.12\columnwidth}\raggedright
R@P$\geq$0.90\strut
\end{minipage} & \begin{minipage}[b]{0.12\columnwidth}\raggedright
Threshold\strut
\end{minipage}\tabularnewline
\midrule
\endhead
\begin{minipage}[t]{0.12\columnwidth}\raggedright
Random Forest\strut
\end{minipage} & \begin{minipage}[t]{0.12\columnwidth}\raggedright
plain\strut
\end{minipage} & \begin{minipage}[t]{0.12\columnwidth}\raggedright
\textbf{0.861 $\pm$ 0.021}\strut
\end{minipage} & \begin{minipage}[t]{0.12\columnwidth}\raggedright
\textbf{0.847}\strut
\end{minipage} & \begin{minipage}[t]{0.12\columnwidth}\raggedright
0.973\strut
\end{minipage} & \begin{minipage}[t]{0.12\columnwidth}\raggedright
\textbf{0.819}\strut
\end{minipage} & \begin{minipage}[t]{0.12\columnwidth}\raggedright
0.500\strut
\end{minipage}\tabularnewline
\begin{minipage}[t]{0.12\columnwidth}\raggedright
Random Forest\strut
\end{minipage} & \begin{minipage}[t]{0.12\columnwidth}\raggedright
tuned threshold\strut
\end{minipage} & \begin{minipage}[t]{0.12\columnwidth}\raggedright
0.859 $\pm$ 0.027\strut
\end{minipage} & \begin{minipage}[t]{0.12\columnwidth}\raggedright
0.847\strut
\end{minipage} & \begin{minipage}[t]{0.12\columnwidth}\raggedright
0.973\strut
\end{minipage} & \begin{minipage}[t]{0.12\columnwidth}\raggedright
0.819\strut
\end{minipage} & \begin{minipage}[t]{0.12\columnwidth}\raggedright
0.420 $\pm$ 0.126\strut
\end{minipage}\tabularnewline
\begin{minipage}[t]{0.12\columnwidth}\raggedright
Gradient Boosting\strut
\end{minipage} & \begin{minipage}[t]{0.12\columnwidth}\raggedright
SMOTE + tuned\strut
\end{minipage} & \begin{minipage}[t]{0.12\columnwidth}\raggedright
0.842 $\pm$ 0.041\strut
\end{minipage} & \begin{minipage}[t]{0.12\columnwidth}\raggedright
0.792\strut
\end{minipage} & \begin{minipage}[t]{0.12\columnwidth}\raggedright
0.972\strut
\end{minipage} & \begin{minipage}[t]{0.12\columnwidth}\raggedright
0.633\strut
\end{minipage} & \begin{minipage}[t]{0.12\columnwidth}\raggedright
0.984 $\pm$ 0.006\strut
\end{minipage}\tabularnewline
\begin{minipage}[t]{0.12\columnwidth}\raggedright
Random Forest\strut
\end{minipage} & \begin{minipage}[t]{0.12\columnwidth}\raggedright
SMOTE + tuned\strut
\end{minipage} & \begin{minipage}[t]{0.12\columnwidth}\raggedright
0.838 $\pm$ 0.028\strut
\end{minipage} & \begin{minipage}[t]{0.12\columnwidth}\raggedright
0.837\strut
\end{minipage} & \begin{minipage}[t]{0.12\columnwidth}\raggedright
0.979\strut
\end{minipage} & \begin{minipage}[t]{0.12\columnwidth}\raggedright
0.776\strut
\end{minipage} & \begin{minipage}[t]{0.12\columnwidth}\raggedright
0.730 $\pm$ 0.040\strut
\end{minipage}\tabularnewline
\begin{minipage}[t]{0.12\columnwidth}\raggedright
MLP\strut
\end{minipage} & \begin{minipage}[t]{0.12\columnwidth}\raggedright
SMOTE + tuned\strut
\end{minipage} & \begin{minipage}[t]{0.12\columnwidth}\raggedright
0.820 $\pm$ 0.022\strut
\end{minipage} & \begin{minipage}[t]{0.12\columnwidth}\raggedright
0.823\strut
\end{minipage} & \begin{minipage}[t]{0.12\columnwidth}\raggedright
0.961\strut
\end{minipage} & \begin{minipage}[t]{0.12\columnwidth}\raggedright
0.740\strut
\end{minipage} & \begin{minipage}[t]{0.12\columnwidth}\raggedright
0.984 $\pm$ 0.009\strut
\end{minipage}\tabularnewline
\begin{minipage}[t]{0.12\columnwidth}\raggedright
Logistic Regression\strut
\end{minipage} & \begin{minipage}[t]{0.12\columnwidth}\raggedright
tuned threshold\strut
\end{minipage} & \begin{minipage}[t]{0.12\columnwidth}\raggedright
0.767 $\pm$ 0.027\strut
\end{minipage} & \begin{minipage}[t]{0.12\columnwidth}\raggedright
0.754\strut
\end{minipage} & \begin{minipage}[t]{0.12\columnwidth}\raggedright
0.973\strut
\end{minipage} & \begin{minipage}[t]{0.12\columnwidth}\raggedright
0.342\strut
\end{minipage} & \begin{minipage}[t]{0.12\columnwidth}\raggedright
0.124 $\pm$ 0.076\strut
\end{minipage}\tabularnewline
\begin{minipage}[t]{0.12\columnwidth}\raggedright
Logistic Regression\strut
\end{minipage} & \begin{minipage}[t]{0.12\columnwidth}\raggedright
plain\strut
\end{minipage} & \begin{minipage}[t]{0.12\columnwidth}\raggedright
0.721 $\pm$ 0.036\strut
\end{minipage} & \begin{minipage}[t]{0.12\columnwidth}\raggedright
0.754\strut
\end{minipage} & \begin{minipage}[t]{0.12\columnwidth}\raggedright
0.973\strut
\end{minipage} & \begin{minipage}[t]{0.12\columnwidth}\raggedright
0.342\strut
\end{minipage} & \begin{minipage}[t]{0.12\columnwidth}\raggedright
0.500\strut
\end{minipage}\tabularnewline
\begin{minipage}[t]{0.12\columnwidth}\raggedright
Gradient Boosting\strut
\end{minipage} & \begin{minipage}[t]{0.12\columnwidth}\raggedright
plain\strut
\end{minipage} & \begin{minipage}[t]{0.12\columnwidth}\raggedright
0.589 $\pm$ 0.064\strut
\end{minipage} & \begin{minipage}[t]{0.12\columnwidth}\raggedright
0.534\strut
\end{minipage} & \begin{minipage}[t]{0.12\columnwidth}\raggedright
0.791\strut
\end{minipage} & \begin{minipage}[t]{0.12\columnwidth}\raggedright
0.000\strut
\end{minipage} & \begin{minipage}[t]{0.12\columnwidth}\raggedright
0.500\strut
\end{minipage}\tabularnewline
\begin{minipage}[t]{0.12\columnwidth}\raggedright
Autoencoder\strut
\end{minipage} & \begin{minipage}[t]{0.12\columnwidth}\raggedright
tuned threshold\strut
\end{minipage} & \begin{minipage}[t]{0.12\columnwidth}\raggedright
0.545 $\pm$ 0.039\strut
\end{minipage} & \begin{minipage}[t]{0.12\columnwidth}\raggedright
0.540\strut
\end{minipage} & \begin{minipage}[t]{0.12\columnwidth}\raggedright
0.947\strut
\end{minipage} & \begin{minipage}[t]{0.12\columnwidth}\raggedright
0.142\strut
\end{minipage} & \begin{minipage}[t]{0.12\columnwidth}\raggedright
0.058 $\pm$ 0.036\strut
\end{minipage}\tabularnewline
\bottomrule
\end{longtable}

Two observations invite a general conclusion. First, the best
configuration in the entire table is the \emph{simplest}: a Random
Forest with no resampling, no class weighting, and the default
threshold. Second, every intervention applied to that Random Forest
makes it worse --- threshold tuning is neutral ($\Delta$F1 = -0.002), class
weighting costs 1.7 F1 points, and SMOTE costs 2.3 points and 1.0 point
of AUPRC.

Reported on its own, this is a tidy and attractive finding: \emph{for a
well-calibrated ensemble on severely imbalanced data, the standard
imbalance-handling toolkit is unnecessary and counterproductive.} It is
precisely the type of claim the single-dataset literature routinely
advances. Section 5 tests it.

\begin{center}\rule{0.5\linewidth}{0.5pt}\end{center}

\hypertarget{study-b-the-conclusion-reverses}{%
\section{5. Study B: The Conclusion
Reverses}\label{study-b-the-conclusion-reverses}}

\hypertarget{the-reversal}{%
\subsection{5.1 The reversal}\label{the-reversal}}

Table 2 contrasts, for each model family, the paired $\Delta$F1 from threshold
tuning on the fraud dataset against the mean across the 45-task suite.

\textbf{Table 2.} $\Delta$F1 from threshold tuning: fraud dataset versus
45-task suite. Suite values are mean $\pm$ standard deviation over 135 runs
per family (45 tasks $\times$ 3 folds).

\begin{longtable}[]{@{}lrrl@{}}
\toprule
Model family & $\Delta$F1, credit-card fraud & $\Delta$F1, 45-task suite &
Agreement\tabularnewline
\midrule
\endhead
Logistic Regression & +0.045 & +0.044 $\pm$ 0.087 &
replicates\tabularnewline
Histogram Gradient Boosting & +0.057 & +0.057 $\pm$ 0.091 &
replicates\tabularnewline
MLP & +0.015 & +0.035 $\pm$ 0.071 & replicates\tabularnewline
\textbf{Random Forest} & \textbf{-0.002} & \textbf{+0.101 $\pm$ 0.134} &
\textbf{reverses}\tabularnewline
\bottomrule
\end{longtable}

Three of four families replicate their fraud-dataset behaviour to within
0.02 F1. Random Forest does not. On the fraud dataset it is the model
for which threshold tuning is worthless; across the suite it is the
family for which threshold tuning is worth the \emph{most}, by a margin
of 44 F1 points over the second-placed family. Figure 1 displays the
contrast.

\textbf{Figure 1.} $\Delta$F1 from threshold tuning by model family:
credit-card fraud alone versus the 45-task suite (mean $\pm$ standard
error). Three families agree; Random Forest reverses.

\includegraphics[width=0.95\linewidth]{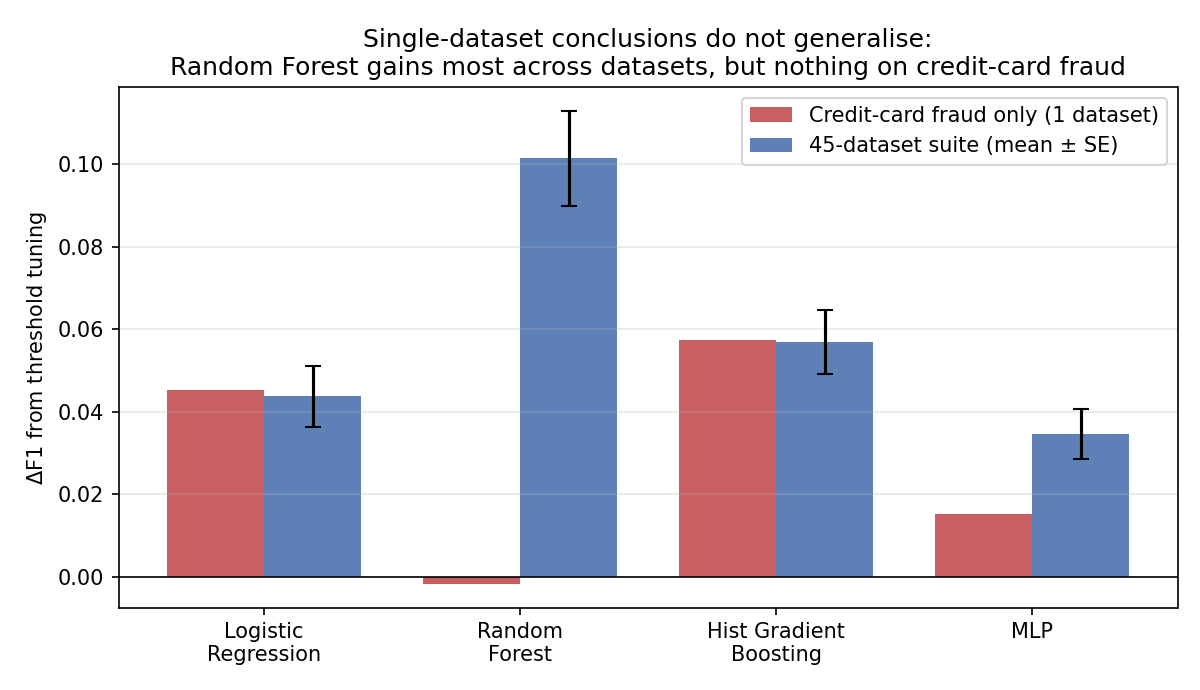}

The consequence for Study A is direct. The finding that made Study A
interesting --- that the best model needs no imbalance handling --- is a
property of Random Forest \emph{on this dataset}, not a property of
Random Forest. A practitioner who read Study A and concluded that Random
Forest can be deployed at the default threshold would, on a task drawn
from our suite, forgo an average of 10 F1 points.

\hypertarget{resampling-reverses-too}{%
\subsection{5.2 Resampling reverses too}\label{resampling-reverses-too}}

The same pattern holds for SMOTE. On the fraud dataset SMOTE reduced
Random Forest F1 by 2.3 points and Logistic Regression F1 by 9.2 points,
motivating the conclusion that SMOTE is harmful for well-behaved models.
Across the 45-task suite, SMOTE is beneficial on average and
significantly so.

\textbf{Table 3.} Strategy effects across the 45-task suite, paired
against the \texttt{plain} configuration at the (dataset, model) level.
\emph{n} = 180 pairs for threshold tuning and SMOTE; 135 for class
weighting (MLP excluded).

\begin{longtable}[]{@{}lrrrrr@{}}
\toprule
Strategy & Mean $\Delta$F1 & Median $\Delta$F1 & Wins & Losses & Wilcoxon
\emph{p}\tabularnewline
\midrule
\endhead
Tuned threshold & +0.0592 & +0.0260 & 138 & 31 & $4.1 \times 10^{-20}$\tabularnewline
Class-balanced + tuned & +0.0547 & +0.0162 & 95 & 34 & $3.2 \times 10^{-09}$\tabularnewline
SMOTE + tuned & +0.0757 & +0.0260 & 138 & 39 & $2.7 \times 10^{-17}$\tabularnewline
\bottomrule
\end{longtable}

The benefit of SMOTE is, however, heavily concentrated by model family:
+0.170 for the MLP, +0.092 for Random Forest, +0.036 for Histogram
Gradient Boosting, and +0.005 --- indistinguishable from nothing --- for
Logistic Regression. A blanket recommendation to resample is therefore
no better founded than a blanket recommendation not to.

\hypertarget{threshold-tuning-benefit-is-non-monotonic-in-imbalance}{%
\subsection{5.3 Threshold-tuning benefit is non-monotonic in
imbalance}\label{threshold-tuning-benefit-is-non-monotonic-in-imbalance}}

Table 4 and Figure 2 group all suite runs by imbalance ratio.

\textbf{Table 4.} $\Delta$F1 from threshold tuning by imbalance ratio band (all
model families pooled).

\begin{longtable}[]{@{}lrrr@{}}
\toprule
Imbalance ratio & Mean $\Delta$F1 & Std & \emph{n} runs\tabularnewline
\midrule
\endhead
\textless{} 1:5 & -0.0088 & 0.039 & 48\tabularnewline
1:5 -- 1:15 & +0.0454 & 0.092 & 192\tabularnewline
\textbf{1:15 -- 1:40} & \textbf{+0.1199} & 0.131 & 84\tabularnewline
1:40 -- 1:100 & +0.0996 & 0.131 & 72\tabularnewline
\textgreater{} 1:100 & +0.0447 & 0.060 & 144\tabularnewline
\bottomrule
\end{longtable}

\textbf{Figure 2.} Threshold-tuning benefit against imbalance ratio,
showing the inverted-U.

\includegraphics[width=0.95\linewidth]{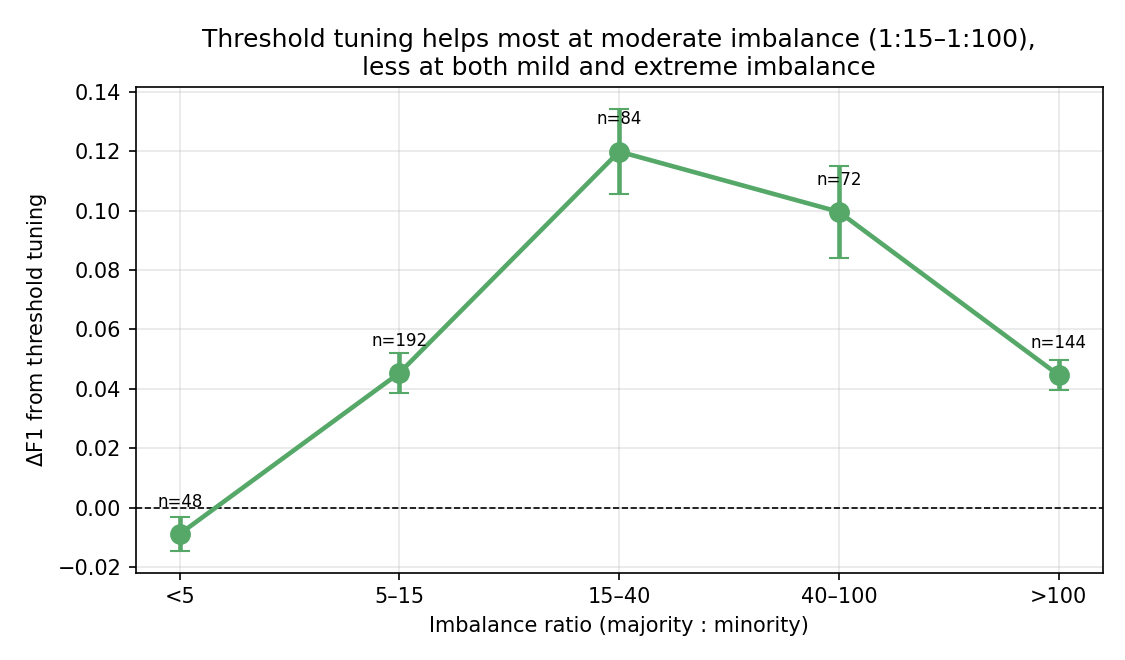}

The relationship is inverted-U rather than monotone. At mild imbalance
the default threshold is already near-optimal and tuning is marginally
harmful. Benefit peaks in the 1:15--1:40 band. Beyond 1:100 it declines
again --- plausibly because at extreme imbalance the validation fold
contains too few positives for the F1-maximising threshold to be
estimated stably, so the selected threshold generalises poorly to the
test fold.

This directly explains Study A. The credit-card fraud dataset sits at
1:577, deep in the regime where threshold tuning delivers least and is
estimated least reliably. It is, on this evidence, an unrepresentative
dataset on which to study threshold selection --- despite being one of
the most popular.

\begin{center}\rule{0.5\linewidth}{0.5pt}\end{center}

\hypertarget{what-predicts-the-benefit-of-threshold-tuning}{%
\section{6. What Predicts the Benefit of Threshold
Tuning?}\label{what-predicts-the-benefit-of-threshold-tuning}}

\hypertarget{a-tempting-heuristic-and-its-failure}{%
\subsection{6.1 A tempting heuristic, and its
failure}\label{a-tempting-heuristic-and-its-failure}}

If threshold tuning corrects for miscalibrated scores, then a model's
calibration error ought to predict how much tuning will help. This would
be practically valuable: calibration error is measurable on the
validation fold before any tuning is attempted, so a practitioner could
tune selectively rather than always. We tested this hypothesis on all
540 paired tuning runs.

\textbf{Table 5.} Correlation between validation-fold calibration
statistics and $\Delta$F1 from threshold tuning (\emph{n} = 540).

\begin{longtable}[]{@{}lrrrr@{}}
\toprule
\begin{minipage}[b]{0.14\columnwidth}\raggedright
Statistic\strut
\end{minipage} & \begin{minipage}[b]{0.18\columnwidth}\raggedleft
Pearson \emph{r}\strut
\end{minipage} & \begin{minipage}[b]{0.18\columnwidth}\raggedleft
\emph{p}\strut
\end{minipage} & \begin{minipage}[b]{0.18\columnwidth}\raggedleft
Spearman $\rho$\strut
\end{minipage} & \begin{minipage}[b]{0.18\columnwidth}\raggedleft
\emph{p}\strut
\end{minipage}\tabularnewline
\midrule
\endhead
\begin{minipage}[t]{0.14\columnwidth}\raggedright
Expected calibration error (equal-width)\strut
\end{minipage} & \begin{minipage}[t]{0.18\columnwidth}\raggedleft
-0.087\strut
\end{minipage} & \begin{minipage}[t]{0.18\columnwidth}\raggedleft
$4.4 \times 10^{-2}$\strut
\end{minipage} & \begin{minipage}[t]{0.18\columnwidth}\raggedleft
-0.075\strut
\end{minipage} & \begin{minipage}[t]{0.18\columnwidth}\raggedleft
$8.1 \times 10^{-2}$\strut
\end{minipage}\tabularnewline
\begin{minipage}[t]{0.14\columnwidth}\raggedright
Expected calibration error (equal-frequency)\strut
\end{minipage} & \begin{minipage}[t]{0.18\columnwidth}\raggedleft
-0.060\strut
\end{minipage} & \begin{minipage}[t]{0.18\columnwidth}\raggedleft
$1.6 \times 10^{-1}$\strut
\end{minipage} & \begin{minipage}[t]{0.18\columnwidth}\raggedleft
+0.032\strut
\end{minipage} & \begin{minipage}[t]{0.18\columnwidth}\raggedleft
$4.5 \times 10^{-1}$\strut
\end{minipage}\tabularnewline
\begin{minipage}[t]{0.14\columnwidth}\raggedright
Brier score\strut
\end{minipage} & \begin{minipage}[t]{0.18\columnwidth}\raggedleft
+0.137\strut
\end{minipage} & \begin{minipage}[t]{0.18\columnwidth}\raggedleft
$1.5 \times 10^{-3}$\strut
\end{minipage} & \begin{minipage}[t]{0.18\columnwidth}\raggedleft
+0.255\strut
\end{minipage} & \begin{minipage}[t]{0.18\columnwidth}\raggedleft
$1.9 \times 10^{-9}$\strut
\end{minipage}\tabularnewline
\bottomrule
\end{longtable}

\textbf{Figure 3.} Calibration statistics against $\Delta$F1. No usable
relationship.

\includegraphics[width=0.95\linewidth]{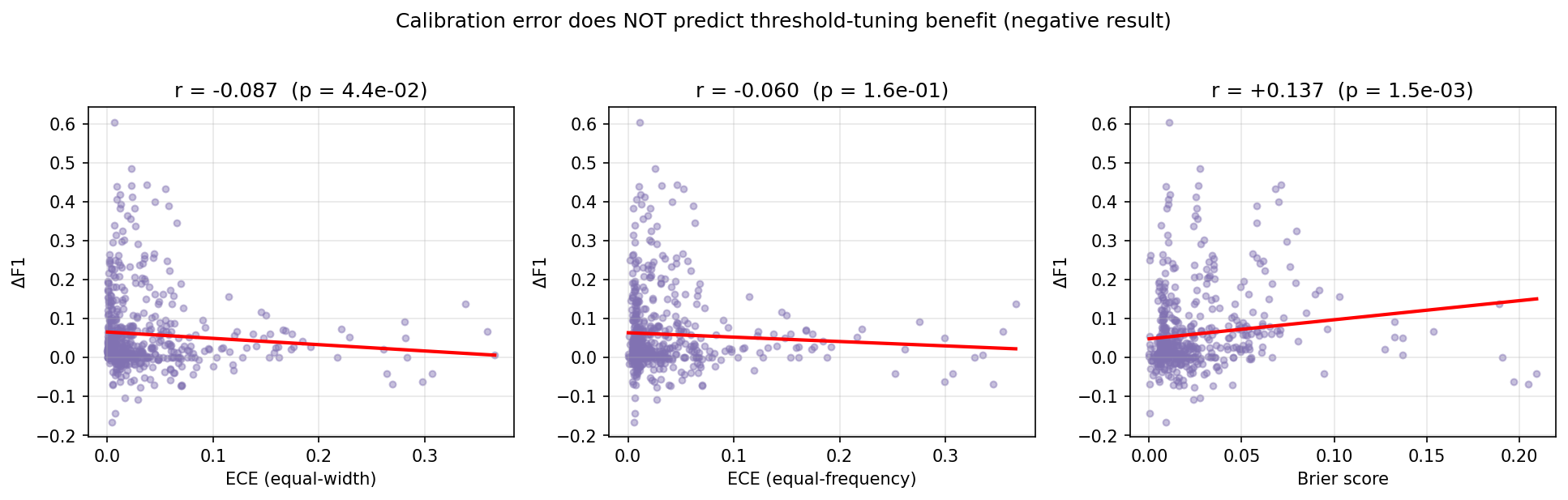}

The hypothesis is rejected. Expected calibration error, in either
binning scheme, is uncorrelated with the benefit of threshold tuning ---
the equal-width variant is in fact very slightly \emph{negatively}
correlated. The Brier score reaches statistical significance but
explains under 2 \% of variance, which is of no practical use for a
go/no-go decision.

The per-family breakdown shows why a global relationship fails to
appear: the sign of the association differs across families. Histogram
Gradient Boosting alone shows a positive within-family association
(\emph{r} = +0.259, \emph{p} = $2.4 \times 10^{-3}$), while Logistic Regression
(\emph{r} = -0.001), MLP (\emph{r} = -0.034), and Random Forest
(\emph{r} = -0.087) show none. Notably, the MLP has by far the largest
mean calibration error (0.086 versus 0.012--0.027 for the others) yet
the \emph{smallest} mean tuning benefit --- the opposite of the
hypothesis.

We report this as a negative result. Aggregate calibration error is not
a usable diagnostic for deciding whether to tune a decision threshold,
and the intuition that it should be is wrong.

\hypertarget{what-does-correlate}{%
\subsection{6.2 What does correlate}\label{what-does-correlate}}

Among the variables we recorded, none is a strong predictor. The log
imbalance ratio (\emph{r} = +0.138) and Brier score (\emph{r} = +0.137)
are weakly positive; AUPRC is weakly negative (\emph{r} = -0.162,
\emph{p} = $1.5 \times 10^{-4}$), consistent with the reading that models which
already rank well have less to gain from moving the operating point.
Sample size and dimensionality are uninformative
(\textbar{}\emph{r}\textbar{} \textless{} 0.09).

The practical implication is unwelcome but clear: there is no cheap
proxy. Because threshold tuning on a validation fold is itself
inexpensive --- a single sweep over a fitted model, requiring no
retraining --- the defensible recommendation is to \emph{always perform
it and let the validation fold decide}, rather than to predict in
advance whether it will pay off. Where it does not help, the validation
fold will select a threshold near 0.5 and little is lost.

\begin{center}\rule{0.5\linewidth}{0.5pt}\end{center}

\hypertarget{discussion}{%
\section{7. Discussion}\label{discussion}}

\hypertarget{limitations}{%
\subsection{7.1 Limitations}\label{limitations}}

\textbf{Suite composition is the principal limitation.} Of the 45 tasks,
30 are synthetic and 15 are real, and the real tasks derive from only
three underlying sources (digits, breast cancer, wine). We were unable
to include established real-world imbalanced benchmarks --- the
imbalanced-learn suite of 27 datasets, OpenML collections, KDD Cup 99,
Covertype --- because the execution environment blocked external data
downloads. The synthetic arm provides controlled variation of the
imbalance ratio that observational data cannot, and we regard it as a
genuine strength for the inverted-U analysis in Section 5.3; but it
cannot substitute for domain diversity when the claim is about external
validity. \textbf{A replication on the imbalanced-learn benchmark suite
is the single most valuable extension of this work}, and the released
harness runs unmodified on it.

\textbf{Maximum imbalance in the suite is 1:178}, well short of the
fraud dataset's 1:577. The declining arm of the inverted-U is therefore
established over a narrower range than we would like, and the
explanation we offer for it (unstable threshold estimation from few
validation positives) is inferred rather than directly demonstrated.

\textbf{Three folds in Study B versus five in Study A}, a compute-budget
concession that widens confidence intervals on the suite estimates.

\textbf{F1 as the selection criterion.} We optimise and report F1
throughout. F1 weights precision and recall equally, which rarely
matches deployment economics. A cost-sensitive criterion with realistic
false-positive and false-negative costs would be more decision-relevant,
and might alter which strategies appear best.

\textbf{No temporal validation.} We use random stratified folds. Hayat
\& Magnier (2025) identify inadequate temporal validation as a major
issue for the fraud dataset specifically, which has a usable
\texttt{Time} axis. Our Study A results inherit this limitation.

\hypertarget{implications}{%
\subsection{7.2 Implications}\label{implications}}

\textbf{For practitioners.} Perform threshold selection on a validation
fold as a matter of routine; it is cheap, it helps on average across
model families and imbalance regimes, and where it does not help it
costs almost nothing. Do not rely on calibration diagnostics to decide
whether to bother. Treat resampling as model-specific: strongly worth
trying for neural networks, marginal for logistic regression.

\textbf{For researchers.} A comparison of imbalance-handling strategies
conducted on one dataset establishes a fact about that dataset. Our
results show that such facts can invert on other data even when the
original experiment is methodologically clean. Claims of the form
``method X is unnecessary for model Y'' require multi-dataset evidence;
we would encourage a minimum of a dozen tasks spanning at least two
orders of magnitude of imbalance ratio.

\textbf{For reviewers.} When a submission draws a general conclusion
about imbalance handling from a single benchmark, the appropriate
question is not only whether the protocol was leakage-free ---
increasingly it is --- but whether the conclusion was tested anywhere
else.

\hypertarget{conclusion}{%
\subsection{7.3 Conclusion}\label{conclusion}}

We set out to establish a clean, leakage-free result on the
most-benchmarked credit-card fraud dataset, and did: a plain Random
Forest at the default threshold attains F1 = 0.861 $\pm$ 0.021, with every
imbalance intervention leaving it unchanged or worse. We then tested
that conclusion on 45 further binary tasks and found it to be an
artifact of the dataset. Random Forest, the family for which threshold
tuning was worthless on fraud data, gains more from threshold tuning
than any other family across the suite; SMOTE, harmful on fraud data, is
beneficial across the suite. Threshold-tuning benefit peaks at moderate
imbalance and declines at the extreme ratios that make fraud detection a
popular benchmark, which explains the discrepancy. Finally, the
intuitive proposal that calibration error should predict tuning benefit
does not survive contact with the data.

The methodological point generalises beyond this application. Internal
validity --- leakage-free protocols, honest threshold selection --- has
rightly received attention in this literature. External validity has
not, and a methodologically impeccable single-dataset finding can still
be, on the point of interest, backwards.

\begin{center}\rule{0.5\linewidth}{0.5pt}\end{center}

\hypertarget{acknowledgements}{%
\section{Acknowledgements}\label{acknowledgements}}

The author thanks Prof.~Raja Jurdak for supervision and feedback that
shaped the direction of this work, and Dr.~Khizar Hayat for arXiv
endorsement. Responsibility for all content, and for any errors, rests
with the author.

\begin{center}\rule{0.5\linewidth}{0.5pt}\end{center}

\hypertarget{references}{%
\section{References}\label{references}}

\begin{enumerate}
\def\labelenumi{\arabic{enumi}.}
\tightlist
\item
  Chawla, N. V., Bowyer, K. W., Hall, L. O., \& Kegelmeyer, W. P.
  (2002). SMOTE: Synthetic Minority Over-sampling Technique.
  \emph{Journal of Artificial Intelligence Research}, 16, 321--357.
\item
  Dal Pozzolo, A. (2015). \emph{Adaptive Machine Learning for Credit
  Card Fraud Detection}. PhD thesis, Université Libre de Bruxelles.
\item
  Dal Pozzolo, A., Caelen, O., Johnson, R. A., \& Bontempi, G. (2015).
  Calibrating probability with undersampling for unbalanced
  classification. \emph{IEEE Symposium Series on Computational
  Intelligence}, 159--166.
\item
  Dal Pozzolo, A., Boracchi, G., Caelen, O., Alippi, C., \& Bontempi, G.
  (2014). Learned lessons in credit card fraud detection from a
  practitioner perspective. \emph{Expert Systems with Applications},
  41(10), 4915--4928.
\item
  Esposito, C., Landrum, G. A., Schneider, N., Stiefl, N., \& Riniker,
  S. (2021). GHOST: Adjusting the decision threshold to handle
  imbalanced data in machine learning. \emph{Journal of Chemical
  Information and Modeling}, 61(6), 2623--2640.
\item
  Hayat, K., \& Magnier, B. (2025). Data leakage and deceptive
  performance: a critical examination of credit card fraud detection
  methodologies. \emph{Mathematics}, 13(16), 2563.
  https://doi.org/10.3390/math13162563 (preprint: arXiv:2506.02703).
\item
  Kabane, S. (2024). Impact of sampling techniques and data leakage on
  XGBoost performance in credit card fraud detection. arXiv:2412.07437.
\item
  Popova, I., \& Gardi, H. A. A. (2025). Credit card fraud detection.
  arXiv:2509.15044.
\item
  Provost, F., \& Fawcett, T. (2001). Robust classification for
  imprecise environments. \emph{Machine Learning}, 42(3), 203--231.
\item
  Sheng, V. S., \& Ling, C. X. (2006). Thresholding for making
  classifiers cost-sensitive. \emph{Proceedings of AAAI}, 476--481.
\item
  Singh, M. T., et al.~(2024). Heterogeneous graph auto-encoder for
  credit-card fraud detection. arXiv:2410.08121.
\item
  \emph{Balancing the scales: a comprehensive study on tackling class
  imbalance in binary classification.} (2024). arXiv:2409.19751.
\item
  \emph{Improving credit card fraud detection through
  transformer-enhanced GAN oversampling.} (2025). arXiv:2509.19032.
\item
  \emph{Enhancing credit card fraud detection using a stacking
  ensemble.} (2024). \emph{International Journal of Advanced Computer
  Science and Applications}, 15(10).
\item
  Pedregosa, F., et al.~(2011). Scikit-learn: machine learning in
  Python. \emph{Journal of Machine Learning Research}, 12, 2825--2830.
\item
  Lemaître, G., Nogueira, F., \& Aridas, C. K. (2017). Imbalanced-learn:
  a Python toolbox to tackle the curse of imbalanced datasets.
  \emph{Journal of Machine Learning Research}, 18(17), 1--5.
\end{enumerate}

\begin{center}\rule{0.5\linewidth}{0.5pt}\end{center}

\hypertarget{reproducibility}{%
\section{Reproducibility}\label{reproducibility}}

All code and per-run metrics are released.
\texttt{code/fraud\_pipeline.py} implements Study A;
\texttt{code/multi\_dataset\_study.py} implements Study B and is
resumable. \texttt{code/results\_nested\_cv/per\_fold\_metrics.csv} (85
rows) and \texttt{code/results\_multi/per\_run\_metrics.csv} (2,025
rows) contain every reported number. Random seeds are fixed throughout.
Study A runs in approximately nine minutes and Study B in approximately
eight minutes on a consumer laptop, with no GPU requirement and no
dependency outside scikit-learn, imbalanced-learn, pandas, NumPy, SciPy,
and Matplotlib.

\end{document}